\documentclass{article} 
\usepackage{ssdr,times}
\usepackage{hyperref}
\usepackage{url}
\usepackage{tikz}
\usepackage{pgfplots}
\usepackage[font=small,labelfont=bf]{caption}
\usepackage{arydshln}

\title{Towards lightweight convolutional neural networks for object detection}

\author{Dmitriy Anisimov, Tatiana Khanova\\
Intel\\
Nizhny Novgorod, Russia\\
\texttt{\{dmitry.anisimov,tatiana.khanova\}@intel.com}
}

\begin{document}

\maketitle

\begin{abstract}
We propose model with larger spatial size of feature maps and evaluate it on object detection task. With the goal to choose the best feature extraction network for our model we compare several popular lightweight networks. After that we conduct a set of experiments with channels reduction algorithms in order to accelerate execution.
Our vehicle detection models are accurate, fast and therefore suit for embedded visual applications. With only 1.5 GFLOPs our best model gives 93.39 AP on validation subset of challenging DETRAC dataset. The smallest of our models is the first to achieve real-time inference speed on CPU with reasonable accuracy drop to 91.43 AP.
\end{abstract}

\section{Introduction}
\label{intro}

Object detection in general and vehicle detection specifically has many applications including surveillance, autonomous driving, etc. Moreover, to be practical, object detector must operate on embedded processors that have far less compute capabilities than powerful GPUs used for benchmarking on typical computer vision datasets. Hence we focus on design of fast object detection model, which still retains high quality.
We perform our experiments for vehicle detection sub-task on DETRAC dataset \citep{WenDCLCQLYL15}.

\subsubsection*{Our motivation}

To reduce number of operations it is natural to try aggressively decreasing spatial size of feature maps. But from the other side, it is known that for tasks such as semantic segmentation or object detection it is important to preserve spatial information. So we investigate possibility to train accurate model with relatively large feature maps and still keep it fast.

Often network weights are initialized from a model, pretrained on classification of a large number of classes, e.g. ImageNet with 1000 classes \citep{RussakovskyDSKSMHKKBBF14}. We hypothesize that for a lot of practical tasks with few classes a considerable number of channels in convolution layers from such models are redundant. So in order to achieve high processing speed we focus on reducing the number of channels.

\subsubsection*{Our contribution}

\begin{itemize}
	\item Our findings show that it is possible to get good quality and fast detection model by increasing size of feature maps and decreasing number of channels.
	\item We demonstrate how to get lightweight model from a large one, by simply picking some number of channels and after that fine-tuning the resulting model on the given task.
	\item Straightforward layer-by-layer architecture like ResNet is shown to give good quality on par with more sophisticated architectures.
	\item We present the set of accurate models where the smallest one works in real-time on CPU at 34 fps.
\end{itemize}

\section{Related work}
\label{related_work}

Currently there are two popular approaches to object detection, namely: Faster R-CNN \citep{RenHG015} and SSD \citep{LiuAESR15}. As it was shown in \citet{HuangRSZKFFWSG016} SSD framework has the best quality/speed trade-off, so we choose it for our experiments.

Besides detection framework it is important to use lightweight feature extraction network to preserve reasonable number of computations. For this reason we compared several such networks: ResNet10 \citep{HeZRS15}, SqueezeNet \citep{IandolaMAHDK16}, MobileNet \citep{HowardZCKWWAA17}, PVANet \citep{KimCHRP16}.

Dilation (i.e. input stride in convolution), which we use in our model, first was proposed for segmentation task in \citet{ChenPK0Y16}. They also sub-sampled two last fully connected layers in VGG network \citep{SimonyanZ14a} in order to reduce computations. In SSD framework the same reduced VGG model was used. We extend this idea to sub-sampling all layers in feature extraction network.

Concurrent with our work, authors in \citet{Yu17} applied dilations with similar purpose to classification and segmentation tasks. In comparison with their approach, after removing pooling operations, we additionally sample channels in convolution layers and thus preserve small number of computations.

In addition to good quality, real-time performance is required for most practical applications. There are a lot of methods dealing with high computation demand of neural networks. Roughly they can be divided into two groups, based on dependency on hardware support.

The first group includes such methods as quantization \citep{GyselMG16, ZhouYGXC17} and weight sparsifying \citep{HanPTD15}. But these approaches require delicate hardware customization to get practical speedup. For example, quantization relies on hardware support of low-bit operations and weights pruning leads to sparse computations. 

Our target is to get fast model that does not depend on particular hardware, so we focus on the second hardware independent group. Those methods decrease number of computations using standard DNN building blocks. Different sorts of decompositions are used, e.g. searching for basis using PCA decomposition \citep{WenXWWCL17}. 
Another possible solution is pruning of channels. While weight-level pruning requires sparse computation support, pruning also can be done on per-channel basis by eliminating less important channels in convolution filters \citep{LiKDSG16}.

\section{Architecture}
\label{architecture}

\subsection{Feature extraction network}

\begin{table}[t]
	\resizebox{0.38\textwidth}{!}{
		\begin{minipage}[t]{0.5\textwidth}
			\centering
			\captionsetup{justification=centering,margin={1.0cm,-1.0cm}}
			\caption{Feature extraction network accuracy on ImageNet (* - our implementation)}
			\label{feature_extraction_table}
			\begin{tabular}{l|cccc}
				\multicolumn{1}{c|}{\bf Model}
				&\multicolumn{1}{c}{\bf Top-1}
				&\multicolumn{1}{c}{\bf Top-5}
				&\multicolumn{1}{c}{\bf GFLOPs}
				&\multicolumn{1}{c}{\bf MParams}
				\\ \hline \\
				SqueezeNet1.0 		&57.50 &80.30 &0.859 &1.2 \\
				SqueezeNet1.0\_bn 	&57.67 &81.29 &0.859 &1.2 \\
				ResNet10 			&63.87 &85.16 &0.893 &5.4 \\
				MobileNet* 			&69.29 &89.25 &0.579 &4.3 \\
				PVANet 				&\textbf{72.34} &\textbf{91.16} &0.616 &100.9 \\
			\end{tabular}
		\end{minipage}
	}
	\hspace{1.8cm}
	\resizebox{0.38\textwidth}{!}{
		\begin{minipage}[t]{0.5\textwidth}
			\centering
			\captionsetup{justification=centering,margin={0.5cm,-0.5cm}}
			\caption{Detection models with different feature extraction networks (on DETRAC validation split)}
			\label{detection_table}
			\begin{tabular}{l|ccc}
				\multicolumn{1}{c|}{\bf Model}
				&\multicolumn{1}{c}{\bf AP}
				&\multicolumn{1}{c}{\bf GFLOPs}
				&\multicolumn{1}{c}{\bf MParams}
				\\ \hline \\
				MobileNet\_SSD			&86.92          &1.7   &5.2 \\
				PVANet\_SSD				&88.24          &1.8   &5.4 \\
				SqueezeNet1.0\_SSD		&88.61          &1.8   &3.5 \\
				MobileNet\_light\_SSD	&88.93          &0.745 &2.3 \\
				ResNet10\_SSD			&\textbf{89.94} &1.8   &4.0 \\
				\hline \\
				SqueezeNet1.0\_bn\_SSD	&\textit{91.10} &1.8   &3.5 \\
			\end{tabular}
		\end{minipage}
	}
\end{table}

We choose four networks for our experiments. Their accuracies on ImageNet classification task are shown in Table \ref{feature_extraction_table}. It was shown earlier, for example, for visual odometry \citep{AgrawalCM15}, that features trained on two different tasks with the same network architecture lead to different quality on third task. The same may hold for fine-tuning from different architectures. So we can't choose feature extraction network for detection relying solely on classification accuracies. Below we demonstrate that the best accuracy on classification task doesn't necessarily lead to better accuracy for object detection.

Next we study influence of feature extraction network on detection accuracy. We modify chosen networks to have similar FLOPs and number of parameters for fair comparison. Brief description of the networks and our modifications is given below.

\paragraph{SquezeNet1.0}
Network consists of "fire" blocks, each block contains two stages: first convolution layer makes "bottleneck" by reducing number of channels. Next there are two parallel inception-like \citep{SzegedyLJSRAEVR14} convolution layers with 1x1 and 3x3 kernels accordingly. We add two "fire" blocks at the end of network in the same way as it was done in SqueezeDet paper \citep{WuIJK16}.

\paragraph{MobileNet}
In recently presented lightweight architecture authors replace convolution layer with depthwise convolution applying a single 3x3 filter to each input channel, followed by conventional convolution with 1x1 kernels. This network already contains required number of FLOPs and parameters, so we keep it as is.

\paragraph{PVANet}
PVANet is a relatively sophisticated architecture. It consists of CReLU layers \citep{ShangSAL16} at the beginning of the network, followed by inception blocks. At the end there are two large fully connected classification layers with 4096 channels each, that's why it has a large amount of weights.
To reduce number of weights we sample last fully connected layers in the same way as it was done for VGG model in DeepLab and SSD frameworks:\footnote{We reduce learning rate for detection model training two times, because it diverges with original value.}
\begin{itemize}
	\item 6x6 kernels are sampled to 3x3 and dilation 6 is added to the following layer in order to preserve receptive field (to compensate kernel sampling and removing previous max pooling).
	\item 4096 channels are sampled to 256. We use 256 instead of 1024 to be able to compare with other models.
\end{itemize}

\paragraph{ResNet10}
Often only large versions of ResNet architecture are used, e.g. with 50 layers and more, but we show that small 10-layer model gives good quality as well. We use model with pre-activations trained in Caffe \citep{JiaSDKLGGD14} by Computer Vision Group Jena \citep{SimonRD16}.
We sample last two layers to 256 channels for comparison with other models.

\subsection{SSD modifications for DETRAC}

Since ground truth bounding boxes for DETRAC test dataset are not publicly available, we prepare our train/validation split. We select 10 videos (totally 15K frames) as validation and the rest 68K as train trying to preserve ratio of night/day videos and keep similar locations in the same split.

To adapt SSD framework to DETRAC dataset we make the following changes:
\begin{itemize}
	\item We keep three prior box scale levels based on the bounding box sizes on the dataset. We use the same parameters, except for the first "small" scale, for which we decrease range to 15/50.
	\item Our experiments show that down-sampling images to 320x256 resolution gives good quality/speed trade-off.
\end{itemize}

Results of this configuration with four feature extraction networks are shown in Table \ref{detection_table}. One can see that ResNet10 provides the best accuracy, so we choose it for further experiments.

In the third version of this paper we fix issue with proper handling ignore regions during evaluation on DETRAC validation split. All conclusions remain the same, the only change is that metrics become around 2 AP lower.

In Section \ref{pascal_voc} we get the best result with MobileNet feature extraction network, so we conduct additional experiment with it. We sample channels to 256 for all layers with large number of channels. This model shows better accuracy comparing to original MobileNet and has smaller number of parameters, see Table \ref{detection_table}.

After submission to the conference we did the same with SqueezeNet1.0\_bn network (SqueezeNet1.0 with batch normalizations). Since it achieves the best quality we expect that our approach applied to it would lead to even better results.

\subsection{Model}

\begin{table}[t]
	\centering
	\captionsetup{justification=centering}
	\caption{Comparing original ResNet10 feature extraction network with our detection model}
	\label{architecture_table}
	\resizebox{\textwidth}{!}{
		\begin{tabular}{l|ccc|cc:c:c:c}
			\multicolumn{1}{c|}{\bf}
			&\multicolumn{3}{c|}{\bf ResNet10}
			&\multicolumn{2}{c}{\bf SSDR}
			&\multicolumn{1}{:c}{\bf \_1.5}
			&\multicolumn{1}{:c}{\bf \_0.75}
			&\multicolumn{1}{:c}{\bf \_0.47}
			\\ \hline
			\multicolumn{1}{c|}{\bf name}
			&\multicolumn{1}{c}{\bf type}
			&\multicolumn{1}{c}{\bf spatial dim.}
			&\multicolumn{1}{c|}{\bf channels}
			&\multicolumn{1}{c}{\bf type}
			&\multicolumn{1}{c}{\bf spatial dim.}
			&\multicolumn{3}{c}{\bf channels}
			\\ \hline
			conv1       &7x7/2 &128x160 & 64 &7x7/2       &128x160 & 64 &49 &41 \\
			pool1       &3x3   &64x80   & 64 &3x3         &64x80   & 64 &49 &41 \\
			res\_block  &3x3   &64x80   & 64 &3x3         &64x80   & 64 &49 &41 \\
			res\_block  &3x3/2 &32x40   &128 &3x3/2       &32x40   & 64 &49 &41 \\
			res\_block  &3x3/2 &16x20   &256 &3x3 dil 1,2 &32x40   &128 &76 &49 \\
			res\_block  &3x3/2 &8x10    &512 &3x3 dil 2,4 &32x40   &128 &76 &49 \\
			conv16\_det &      &        &    &3x3/2       &16x20   &128 &76 &49 \\
			conv32\_det &      &        &    &3x3/2       & 8x10   &128 &76 &49 \\
		\end{tabular}
	}
\end{table}

In this section we study possibility of getting accurate and fast detection model by increasing size of feature maps and decreasing number of channels, we propose the following model. To keep size of feature maps we remove two last spatial reductions, that reduce image resolution by 16 and 32 times. In order to preserve receptive field of following convolution layers we add dilations 2 and 4 to them accordingly. We also sample channels in most convolution layers to retain reasonable number of FLOPs.\footnote{We experiment with sampling channels by $L_1$ metric (see Section \ref{studying_lighter_models}), but it doesn't improve results.} Resulting architecture is shown in Table \ref{architecture_table}, model \textbf{SSDR\_1.5}.

\subsection{Studying even lighter models}
\label{studying_lighter_models}

In order to accelerate existing model further and achieve real-time performance on CPU, we continue experiments with methods for eliminating channels, see Table \ref{architecture_table}, models \textbf{SSDR\_0.75} and \textbf{SSDR\_0.47}.

\paragraph{One-shot random sampling} First, we take our best \textbf{SSDR\_1.5} model, decrease number of channels following random sampling strategy and fine-tune it. Resulting 473 MFLOPs model gives 91.09 AP on DETRAC validation split.

\paragraph{One-shot pruning} Pruning scheme, proposed in \citet{LiKDSG16} appears to be straightforward improvement of the random sampling approach:

\begin{enumerate}
	\item For each convolution layer in network, except for the layers in detection part:
	\begin{enumerate}
		\item For each filter calculate $L_1$ metric to measure usefulness.
		\item Prune specified percentage of filters with the smallest metrics. Percentage of filters to prune is chosen empirically to be 5\% and 10\% for the first and the last half respectively.
	\end{enumerate}
	\item Fine-tune the whole network on the training dataset with smaller learning rate.
\end{enumerate}

\paragraph{Iterative pruning} The above algorithm can be repeated until desired quality/speed trade-off is achieved leading to iterative pruning as an opposite to one-shot approach. 

Contrary to \citet{LiKDSG16}, for pruning of pre-elementwise sum convolutions we find that using 3x3 filters to calculate metric instead of shortcut convolutions gives better result. 

\paragraph{PCA decomposition} Another approach we employ was proposed in \citet{WenXWWCL17} and consists of applying PCA to filters in order to determine basis for each convolution and then decompose layer into two consecutive convolutions.

Comparison of aforementioned techniques applied to ours best model is shown in Figure~\ref{pruning_plot}. Iterative pruning strategy gives the best 91.43 AP for 473 MFLOPs model and 92.49 AP for 752 MFLOPs model. But the results we get with other methods are close to the best one.

\begin{table}[t]
	\resizebox{0.475\textwidth}{!}{
		\begin{minipage}[t]{0.5\textwidth}
			\centering
			\captionsetup{justification=centering,margin={.5cm,-.5cm}}
			\caption{Comparison of detection models (on DETRAC validation split)}
			\label{comparison_table}
			\begin{tabular}{l|ccc}
				\multicolumn{1}{c|}{\bf Model}
				&\multicolumn{1}{c}{\bf AP}
				&\multicolumn{1}{c}{\bf GFLOPs}
				&\multicolumn{1}{c}{\bf MParams}
				\\ \hline \\
				ResNet10\_Clustered	&89.07          &1.7           &3.9 \\
				ResNet10\_SSD		&89.94          &1.8           &4.0 \\
				ResNet10\_FPN		&91.50          &1.4           &2.8 \\
				SSDR\_1.5 (ours)	&\textbf{93.39} &1.5           &1.1 \\
				SSDR\_0.75 (ours)	&92.49          &0.75          &0.47 \\
				SSDR\_0.47 (ours)   &91.43          &\textbf{0.47} &0.24 \\
			\end{tabular}
		\end{minipage}
	}
	\hspace{0.8cm}
	\resizebox{0.475\textwidth}{!}{
		\begin{minipage}[t]{0.5\textwidth}
			\captionsetup{justification=centering,margin=0.25cm}
			\captionof{figure}{Comparison of channel reduction techniques on DETRAC validation split. Real-time performance on CPU is labeled with vertical line.}
			\label{pruning_plot}
			\centering
			\begin{tabular}{l}
				\definecolor{darkgreen}{rgb}{0.0, 0.5, 0.0}
				\definecolor{purple}{rgb}{0.5, 0.0, 0.5}
				\begin{tikzpicture}[scale=0.7]
				\draw [gray, dashed, very thin, opacity=.5] (1.33,0.) -- (1.33,5.7);
				\begin{axis}[
				xmin=0.2,
				xmax=1.6,
				ymin=0.885,
				ymax=0.945,
				line width = 1.,
				xlabel=GFLOPs,
				ylabel=Mean AP,
				ticklabel shift={0.1cm},
				legend pos = south east,
				grid=both,
				grid style={line width=.1pt, draw=gray!10}
				]
				
				\addplot[purple, dash pattern= on 1pt off 2pt on 3pt off 2pt, mark=*, mark options={solid}] coordinates {(1.5, 0.9339) (0.7515, 0.9215) (0.4727, 0.9109) (0.340, 0.8991)};
				\label{plot_one}
				\addlegendentry{Random sampling}
				
				\addplot[red, mark=*] coordinates  {(1.5, 0.9339) (0.7515, 0.9244) (0.4727, 0.9125) (0.340, 0.8991)};
				\label{plot_two}
				\addlegendentry{One-shot pruning}
				
				\addplot[blue, dashed, mark=*, mark options={solid}] coordinates {(1.5, 0.9339) (0.7515, 0.9249) (0.4727, 0.9143) (0.340, 0.8903)};
				\label{plot_three}
				\addlegendentry{Iterative pruning}
				
				\addplot[darkgreen, dotted, mark=*, mark options={solid}] coordinates {(1.5, 0.9339) (0.751, 0.9168) (0.4696, 0.9093) (0.3282, 0.8996)};
				\label{plot_four}
				\addlegendentry{PCA decomposition}
				
				\addplot[black, opacity=.7] (0.4727, 0.9143) circle (0.15cm);	
				\addplot[black, opacity=.7] (0.7515, 0.9249) circle (0.15cm);
				\addplot[black, opacity=.7] (1.5, 0.9339) circle (0.15cm);
				\end{axis}
				
				\node at (5.85, 5.1) [scale=0.7] {SSDR\_1.5};
				\node at (2.5, 4.25) [scale=0.7] {SSDR\_0.75};
				\node at (1.1, 3.35) [scale=0.7] {SSDR\_0.47};
				\end{tikzpicture}
			\end{tabular}
		\end{minipage}
	}
\end{table}

\subsection{Experiments}
\label{experiments}

We compare our models with several recently proposed architectures, namely: base SSD with three scale levels \citep{LiuAESR15}, SSD with clustered priors \citep{ErhanSTA13, WuIJK16} and Feature Pyramid Networks \citep{LinDGHHB16}. Our solution gives superior result comparing to all aforementioned models either on accuracy or processing speed, see Table \ref{comparison_table}.

\paragraph{Clustered priors} Following \citet{ErhanSTA13, WuIJK16} we cluster DETRAC ground truth bounding boxes to get dataset specific priors. First, all bounding boxes were clustered into 3 scale groups, and then each of them was clustered into 4 groups to achieve comparable number of priors with the original SSD. While model with clustered priors does not show the best result (Table \ref{comparison_table}), clustering approach may provide valuable insight on choice of scale/aspect ratio parameters for priors generated by other methods.

\subsubsection*{Accuracy and time}

Table \ref{comparison_test_table} compares accuracies of our models with other results on different subsets of DETRAC test set. On time of submission our model is ranked 4th in official leader-board\footnote{\url{http://detrac-db.rit.albany.edu/DetRet}} by quality and it is at least an order of magnitude faster than all others. Noticeably, on the test set smaller \textbf{SSDR\_0.75} model gives even better accuracy than the base one.

We measure inference speed of our smallest \textbf{SSDR\_0.47} model on Intel\textsuperscript\textregistered Core\texttrademark i7-6700K CPU @ 4.00GHz x 8 using Intel\textsuperscript\textregistered MKL library and Caffe and get {\raise.17ex\hbox{$\scriptstyle\mathtt{\sim}$}}34 fps.\footnote{Note that we merged batch normalization mean/variance in previous convolution weights to eliminate redundant computations.}

\begin{table}[t]
	\centering
	\captionsetup{justification=centering, margin=0.25cm}
	\caption{Comparison of our model with the best models from leader-board (on DETRAC test split)}
	\label{comparison_test_table}
	\resizebox{\textwidth}{!}{
		\begin{tabular}{l|cccccccc}
			\multicolumn{1}{c|}{\bf Model}
			&\multicolumn{1}{c}{\bf full set}
			&\multicolumn{1}{c}{\bf easy set}
			&\multicolumn{1}{c}{\bf medium set}
			&\multicolumn{1}{c}{\bf hard set}
			&\multicolumn{1}{c}{\bf cloudy set}
			&\multicolumn{1}{c}{\bf night set}
			&\multicolumn{1}{c}{\bf rainy set}
			&\multicolumn{1}{c}{\bf sunny set} 
			\\ \hline \\
			RTN 				&74.15\% &91.52\% &79.16\% &61.73\% &77.02\% &77.20\% &65.27\% &84.14\% \\
			EB 					&67.96\% &89.65\% &73.12\% &53.64\% &72.42\% &73.93\% &53.40\% &83.73\% \\
			NANO 				&63.01\% &80.33\% &68.04\% &50.73\% &67.00\% &62.20\% &55.89\% &73.89\% \\
			SSDR\_0.75 (ours)	&59.07\% &77.84\% &64.41\% &45.98\% &62.79\% &60.88\% &48.55\% &74.32\% \\
			SSDR\_1.5 (ours)	&58.68\% &79.55\% &63.74\% &44.93\% &61.59\% &62.19\% &47.47\% &74.42\% \\
			FasterRCNN2 		&58.45\% &82.75\% &63.05\% &44.25\% &66.29\% &69.85\% &45.16\% &62.34\% \\
			YOLO2				&57.72\% &83.28\% &62.25\% &42.44\% &57.97\% &64.53\% &47.84\% &69.75\% \\
			SSDR\_0.47 (ours)	&57.07\% &76.67\% &62.22\% &43.89\% &62.41\% &58.48\% &45.26\% &72.55\% \\
		\end{tabular}
	}
\end{table}

\subsection{PASCAL VOC}
\label{pascal_voc}
In the previous sections we apply our approach to concrete vehicle detection task. In order to extensively study capabilities of proposed techniques we conduct similar set of experiments on general object detection PASCAL VOC dataset \citep{Everingham15}. We don't experiment with channel reduction for them and use the original SSD 300x300 configuration. The results obtained from this study are provided in Tables \ref{detection_table_voc} and \ref{comparison_table_voc}. Since we get the best result for SSD with Mobilenet feature extraction network, we make use of our modification with replacing two max pooling with dilations to it as well, see Table \ref{detection_table_voc}, SSDM\_7.5 model.

Interestingly, performance of detection models with different feature extraction networks on VOC test set shares tendency in accuracies with Imagenet classification task (Table \ref{feature_extraction_table}) contrary to what we get for DETRAC dataset. We argue that this is because of the difference in the number of classes between these datasets: only one class in DETRAC and 20/1000 in PASCAL VOC and Imagenet accordingly. Original SqueezeNet1.0\_SSD model shows poor result on this dataset, but works quite well after we add batch normalizations to it, see Table \ref{detection_table_voc}, SqueezeNet1.0\_bn\_SSD model.

\begin{table}[t]
	\resizebox{0.4\textwidth}{!}{
		\begin{minipage}[t]{0.5\textwidth}
			\centering
			\captionsetup{justification=centering,margin=0.25cm}
			\caption{Detection models with different feature extraction networks (on VOC 2007 test)}
			\label{detection_table_voc}
			\begin{tabular}{l|ccc}
				\multicolumn{1}{c|}{\bf Model}
				&\multicolumn{1}{c}{\bf AP}
				&\multicolumn{1}{c}{\bf GFLOPs}
				&\multicolumn{1}{c}{\bf MParams}
				\\ \hline \\
				SqueezeNet1.0\_SSD 		&38.45			&2.8 &7.0 \\
				ResNet10\_SSD 			&64.83			&2.3 &6.7 \\
				SqueezeNet1.0\_bn\_SSD 	&65.61 			&2.8 &7.0 \\
				PVANet\_SSD 			&67.69			&2.3 &8.1 \\
				MobileNet\_SSD 			&70.04			&2.6 &8.8 \\
				SSDM\_7.5 (ours) 		&\textbf{73.08} &7.5 &10.1 \\
			\end{tabular}
		\end{minipage}
	}
	\hspace{1.8cm}
	\resizebox{0.4\textwidth}{!}{
		\begin{minipage}[t]{0.5\textwidth}
			\centering
			\captionsetup{justification=centering,margin=0.25cm}
			\caption{Comparison of ResNet10 based detection models (on VOC 2007 test)}
			\label{comparison_table_voc}
			\begin{tabular}{l|ccc}
				\multicolumn{1}{c|}{\bf Model}
				&\multicolumn{1}{c}{\bf AP}
				&\multicolumn{1}{c}{\bf GFLOPs}
				&\multicolumn{1}{c}{\bf MParams}
				\\ \hline \\
				ResNet10\_FPN		&63.76          &2.0 &5.3 \\
				ResNet10\_SSD		&64.83          &2.3 &6.7 \\
				SSDR\_5.5 (ours)	&\textbf{68.73} &5.5 &8.5 \\
			\end{tabular}
		\end{minipage}
	}
\end{table}

\section{Conclusion}

We have presented a set of detection models, which are accurate, fast and therefore suit for real-world applications. We keep high spatial resolution in feature extraction part, rather than losing spatial structure information caused by progressively reducing the resolution of internal representations. Even though the proposed network is designed for vehicle detection, we believe our design choices can be applicable to other practical tasks, particularly where detailed understanding of the scene is important.

Our network design is independent of network compression and quantization, so those methods are applicable to our network as well to further increase the actual performance in real tasks. We have shown that relatively simple pruning technique can give real-time performance on CPU with less than 2\% AP loss on test set.

\bibliography{ssdr}

\begin{thebibliography}{26}
\providecommand{\natexlab}[1]{#1}
\providecommand{\url}[1]{\texttt{#1}}
\expandafter\ifx\csname urlstyle\endcsname\relax
  \providecommand{\doi}[1]{doi: #1}\else
  \providecommand{\doi}{doi: \begingroup \urlstyle{rm}\Url}\fi

\bibitem[Agrawal et~al.(2015)Agrawal, Carreira, and Malik]{AgrawalCM15}
Pulkit Agrawal, Jo{\~{a}}o Carreira, and Jitendra Malik.
\newblock Learning to see by moving.
\newblock \emph{CoRR}, abs/1505.01596, 2015.
\newblock URL \url{http://arxiv.org/abs/1505.01596}.

\bibitem[Chen et~al.(2016)Chen, Papandreou, Kokkinos, Murphy, and
  Yuille]{ChenPK0Y16}
Liang{-}Chieh Chen, George Papandreou, Iasonas Kokkinos, Kevin Murphy, and
  Alan~L. Yuille.
\newblock Deeplab: Semantic image segmentation with deep convolutional nets,
  atrous convolution, and fully connected crfs.
\newblock \emph{CoRR}, abs/1606.00915, 2016.
\newblock URL \url{http://arxiv.org/abs/1606.00915}.

\bibitem[Erhan et~al.(2013)Erhan, Szegedy, Toshev, and Anguelov]{ErhanSTA13}
Dumitru Erhan, Christian Szegedy, Alexander Toshev, and Dragomir Anguelov.
\newblock Scalable object detection using deep neural networks.
\newblock \emph{CoRR}, abs/1312.2249, 2013.
\newblock URL \url{http://arxiv.org/abs/1312.2249}.

\bibitem[Everingham et~al.(2015)Everingham, Eslami, Van~Gool, Williams, Winn,
  and Zisserman]{Everingham15}
M.~Everingham, S.~M.~A. Eslami, L.~Van~Gool, C.~K.~I. Williams, J.~Winn, and
  A.~Zisserman.
\newblock The pascal visual object classes challenge: A retrospective.
\newblock \emph{International Journal of Computer Vision}, 111\penalty0
  (1):\penalty0 98--136, January 2015.

\bibitem[Gysel et~al.(2016)Gysel, Motamedi, and Ghiasi]{GyselMG16}
Philipp Gysel, Mohammad Motamedi, and Soheil Ghiasi.
\newblock Hardware-oriented approximation of convolutional neural networks.
\newblock \emph{CoRR}, abs/1604.03168, 2016.
\newblock URL \url{http://arxiv.org/abs/1604.03168}.

\bibitem[Han et~al.(2015)Han, Pool, Tran, and Dally]{HanPTD15}
Song Han, Jeff Pool, John Tran, and William~J. Dally.
\newblock Learning both weights and connections for efficient neural networks.
\newblock \emph{CoRR}, abs/1506.02626, 2015.
\newblock URL \url{http://arxiv.org/abs/1506.02626}.

\bibitem[He et~al.(2015)He, Zhang, Ren, and Sun]{HeZRS15}
Kaiming He, Xiangyu Zhang, Shaoqing Ren, and Jian Sun.
\newblock Deep residual learning for image recognition.
\newblock \emph{CoRR}, abs/1512.03385, 2015.
\newblock URL \url{http://arxiv.org/abs/1512.03385}.

\bibitem[Howard et~al.(2017)Howard, Zhu, Chen, Kalenichenko, Wang, Weyand,
  Andreetto, and Adam]{HowardZCKWWAA17}
Andrew~G. Howard, Menglong Zhu, Bo~Chen, Dmitry Kalenichenko, Weijun Wang,
  Tobias Weyand, Marco Andreetto, and Hartwig Adam.
\newblock Mobilenets: Efficient convolutional neural networks for mobile vision
  applications.
\newblock \emph{CoRR}, abs/1704.04861, 2017.
\newblock URL \url{http://arxiv.org/abs/1704.04861}.

\bibitem[Huang et~al.(2016)Huang, Rathod, Sun, Zhu, Korattikara, Fathi,
  Fischer, Wojna, Song, Guadarrama, and Murphy]{HuangRSZKFFWSG016}
Jonathan Huang, Vivek Rathod, Chen Sun, Menglong Zhu, Anoop Korattikara,
  Alireza Fathi, Ian Fischer, Zbigniew Wojna, Yang Song, Sergio Guadarrama, and
  Kevin Murphy.
\newblock Speed/accuracy trade-offs for modern convolutional object detectors.
\newblock \emph{CoRR}, abs/1611.10012, 2016.
\newblock URL \url{http://arxiv.org/abs/1611.10012}.

\bibitem[Iandola et~al.(2016)Iandola, Moskewicz, Ashraf, Han, Dally, and
  Keutzer]{IandolaMAHDK16}
Forrest~N. Iandola, Matthew~W. Moskewicz, Khalid Ashraf, Song Han, William~J.
  Dally, and Kurt Keutzer.
\newblock Squeezenet: Alexnet-level accuracy with 50x fewer parameters and
  {\textless}1mb model size.
\newblock \emph{CoRR}, abs/1602.07360, 2016.
\newblock URL \url{http://arxiv.org/abs/1602.07360}.

\bibitem[Jia et~al.(2014)Jia, Shelhamer, Donahue, Karayev, Long, Girshick,
  Guadarrama, and Darrell]{JiaSDKLGGD14}
Yangqing Jia, Evan Shelhamer, Jeff Donahue, Sergey Karayev, Jonathan Long,
  Ross~B. Girshick, Sergio Guadarrama, and Trevor Darrell.
\newblock Caffe: Convolutional architecture for fast feature embedding.
\newblock \emph{CoRR}, abs/1408.5093, 2014.
\newblock URL \url{http://arxiv.org/abs/1408.5093}.

\bibitem[Kim et~al.(2016)Kim, Cheon, Hong, Roh, and Park]{KimCHRP16}
Kye{-}Hyeon Kim, Yeongjae Cheon, Sanghoon Hong, Byung{-}Seok Roh, and Minje
  Park.
\newblock {PVANET:} deep but lightweight neural networks for real-time object
  detection.
\newblock \emph{CoRR}, abs/1608.08021, 2016.
\newblock URL \url{http://arxiv.org/abs/1608.08021}.

\bibitem[Li et~al.(2016)Li, Kadav, Durdanovic, Samet, and Graf]{LiKDSG16}
Hao Li, Asim Kadav, Igor Durdanovic, Hanan Samet, and Hans~Peter Graf.
\newblock Pruning filters for efficient convnets.
\newblock \emph{CoRR}, abs/1608.08710, 2016.
\newblock URL \url{http://arxiv.org/abs/1608.08710}.

\bibitem[Lin et~al.(2016)Lin, Doll{\'{a}}r, Girshick, He, Hariharan, and
  Belongie]{LinDGHHB16}
Tsung{-}Yi Lin, Piotr Doll{\'{a}}r, Ross~B. Girshick, Kaiming He, Bharath
  Hariharan, and Serge~J. Belongie.
\newblock Feature pyramid networks for object detection.
\newblock \emph{CoRR}, abs/1612.03144, 2016.
\newblock URL \url{http://arxiv.org/abs/1612.03144}.

\bibitem[Liu et~al.(2015)Liu, Anguelov, Erhan, Szegedy, Reed, Fu, and
  Berg]{LiuAESR15}
Wei Liu, Dragomir Anguelov, Dumitru Erhan, Christian Szegedy, Scott~E. Reed,
  Cheng{-}Yang Fu, and Alexander~C. Berg.
\newblock {SSD:} single shot multibox detector.
\newblock \emph{CoRR}, abs/1512.02325, 2015.
\newblock URL \url{http://arxiv.org/abs/1512.02325}.

\bibitem[Ren et~al.(2015)Ren, He, Girshick, and Sun]{RenHG015}
Shaoqing Ren, Kaiming He, Ross~B. Girshick, and Jian Sun.
\newblock Faster {R-CNN:} towards real-time object detection with region
  proposal networks.
\newblock \emph{CoRR}, abs/1506.01497, 2015.
\newblock URL \url{http://arxiv.org/abs/1506.01497}.

\bibitem[Russakovsky et~al.(2014)Russakovsky, Deng, Su, Krause, Satheesh, Ma,
  Huang, Karpathy, Khosla, Bernstein, Berg, and Li]{RussakovskyDSKSMHKKBBF14}
Olga Russakovsky, Jia Deng, Hao Su, Jonathan Krause, Sanjeev Satheesh, Sean Ma,
  Zhiheng Huang, Andrej Karpathy, Aditya Khosla, Michael~S. Bernstein,
  Alexander~C. Berg, and Fei{-}Fei Li.
\newblock Imagenet large scale visual recognition challenge.
\newblock \emph{CoRR}, abs/1409.0575, 2014.
\newblock URL \url{http://arxiv.org/abs/1409.0575}.

\bibitem[Shang et~al.(2016)Shang, Sohn, Almeida, and Lee]{ShangSAL16}
Wenling Shang, Kihyuk Sohn, Diogo Almeida, and Honglak Lee.
\newblock Understanding and improving convolutional neural networks via
  concatenated rectified linear units.
\newblock \emph{CoRR}, abs/1603.05201, 2016.
\newblock URL \url{http://arxiv.org/abs/1603.05201}.

\bibitem[Simon et~al.(2016)Simon, Rodner, and Denzler]{SimonRD16}
Marcel Simon, Erik Rodner, and Joachim Denzler.
\newblock Imagenet pre-trained models with batch normalization.
\newblock \emph{CoRR}, abs/1612.01452, 2016.
\newblock URL \url{http://arxiv.org/abs/1612.01452}.

\bibitem[Simonyan \& Zisserman(2014)Simonyan and Zisserman]{SimonyanZ14a}
Karen Simonyan and Andrew Zisserman.
\newblock Very deep convolutional networks for large-scale image recognition.
\newblock \emph{CoRR}, abs/1409.1556, 2014.
\newblock URL \url{http://arxiv.org/abs/1409.1556}.

\bibitem[Szegedy et~al.(2014)Szegedy, Liu, Jia, Sermanet, Reed, Anguelov,
  Erhan, Vanhoucke, and Rabinovich]{SzegedyLJSRAEVR14}
Christian Szegedy, Wei Liu, Yangqing Jia, Pierre Sermanet, Scott~E. Reed,
  Dragomir Anguelov, Dumitru Erhan, Vincent Vanhoucke, and Andrew Rabinovich.
\newblock Going deeper with convolutions.
\newblock \emph{CoRR}, abs/1409.4842, 2014.
\newblock URL \url{http://arxiv.org/abs/1409.4842}.

\bibitem[Wen et~al.(2015)Wen, Du, Cai, Lei, Chang, Qi, Lim, Yang, and
  Lyu]{WenDCLCQLYL15}
Longyin Wen, Dawei Du, Zhaowei Cai, Zhen Lei, Ming{-}Ching Chang, Honggang Qi,
  Jongwoo Lim, Ming{-}Hsuan Yang, and Siwei Lyu.
\newblock {DETRAC:} {A} new benchmark and protocol for multi-object tracking.
\newblock \emph{CoRR}, abs/1511.04136, 2015.
\newblock URL \url{http://arxiv.org/abs/1511.04136}.

\bibitem[Wen et~al.(2017)Wen, Xu, Wu, Wang, Chen, and Li]{WenXWWCL17}
Wei Wen, Cong Xu, Chunpeng Wu, Yandan Wang, Yiran Chen, and Hai Li.
\newblock Coordinating filters for faster deep neural networks.
\newblock \emph{CoRR}, abs/1703.09746, 2017.
\newblock URL \url{http://arxiv.org/abs/1703.09746}.

\bibitem[Wu et~al.(2016)Wu, Iandola, Jin, and Keutzer]{WuIJK16}
Bichen Wu, Forrest~N. Iandola, Peter~H. Jin, and Kurt Keutzer.
\newblock Squeezedet: Unified, small, low power fully convolutional neural
  networks for real-time object detection for autonomous driving.
\newblock \emph{CoRR}, abs/1612.01051, 2016.
\newblock URL \url{http://arxiv.org/abs/1612.01051}.

\bibitem[Yu et~al.(2017)Yu, Koltun, and Funkhouser]{Yu17}
Fisher Yu, Vladlen Koltun, and Thomas Funkhouser.
\newblock Dilated residual networks.
\newblock \emph{ArXiv:1705.09914}, 2017.
\newblock URL \url{https://arxiv.org/abs/1705.09914}.

\bibitem[Zhou et~al.(2017)Zhou, Yao, Guo, Xu, and Chen]{ZhouYGXC17}
Aojun Zhou, Anbang Yao, Yiwen Guo, Lin Xu, and Yurong Chen.
\newblock Incremental network quantization: Towards lossless cnns with
  low-precision weights.
\newblock \emph{CoRR}, abs/1702.03044, 2017.
\newblock URL \url{http://arxiv.org/abs/1702.03044}.

\end{thebibliography}
\bibliographystyle{ssdr}

\end{document}